\documentclass[a4paper]{article}

\usepackage[english]{babel}
\usepackage[utf8x]{inputenc}
\usepackage[T1]{fontenc}

\usepackage{geometry}
\usepackage{amsmath,amsfonts,amsthm,amssymb}
\usepackage{bm}
\usepackage{graphicx}
\graphicspath{{figures/}}
\usepackage[colorinlistoftodos]{todonotes}
\usepackage[colorlinks=true, allcolors=blue]{hyperref}
\usepackage{times}
\usepackage{latexsym}
\usepackage{subfigure}
\usepackage{xspace}
\usepackage{multirow}
\usepackage{color}
\usepackage{helvet}  
\usepackage{courier}  
\usepackage{url}  
\usepackage{graphicx}  
\usepackage{named}
\newcommand{\method}{\xspace{SAM}}

\title{\method: Semantic Attribute Modulation for \\Language Modeling and Style Variation}
\author{Wenbo Hu$^1$, Lifeng Hua$^2$, Lei Li$^2$, Hang Su$^2$, Tian Wang$^3$, Ning Chen$^1$, Bo Zhang$^1$\\
$^1$~Tsinghua University\\ 
\{hwb13@mails.,suhangss,ningchen,dcszb\}@tsinghua.edu.cn,\\
$^2$~Toutiao AI LAB, \{lileilab,hualifeng\}@toutiao.com,\\
$^3$~eBay Inc, twang5@ebay.com}
\begin{document}
\maketitle

\begin{abstract}
This paper presents a Semantic Attribute Modulation (SAM) for language modeling and style variation. The semantic attribute modulation includes various document attributes, such as titles, authors, and document categories.
We consider two types of attributes, (title attributes and category attributes), and a \emph{flexible} attribute selection scheme by automatically scoring them via an attribute attention mechanism. 
The semantic attributes are embedded into the hidden semantic space as the generation inputs.
With the attributes properly harnessed, our proposed SAM can generate \emph{interpretable} texts with regard to the input attributes. 
Qualitative analysis, including word semantic analysis and attention values, shows the interpretability of \method. On several typical text datasets, we empirically demonstrate the superiority of the Semantic Attribute Modulated language model with different combinations of document attributes.
Moreover, we present a style variation for the lyric generation using \method, which shows a strong connection between the style variation and the semantic attributes.
\end{abstract}

\section{Introduction}\label{sec:introduction}

Language generation is considered as a key task in the artificial intelligence field~\cite{reiter2000building}.
The language modeling task aims to present the word distributions of text sequences and is considered as a degenerated text generation task, which generates only one word at each step.
Traditional language generation approaches use phrase templates and related generation rules. For the language modeling task, the counting-based \emph{n}-gram method is broadly used. These methods are conceptually simple but hard to generalize like humans.

Later on, Bengio et al.~\shortcite{bengio2003neural} developed a feed-forward neural network language model and Mikolov et al.~\cite{mikolov2010recurrent} used the recurrent neural network~(RNN) to train a language model.
With the benefits of the large-scale corpora and the modified gating functions, such as the long-short term memory (LSTM)~\cite{hochreiter1997long} or the gated recurrent unit (GRU)~\cite{chung2014empirical}, the recurrent neural network (RNN) has been demonstrated a good capability in modeling word probabilities and now is the most widely used method for language modeling and language generation~\cite{mikolov2010recurrent,graves2013generating}.
Nevertheless, RNN is often criticized for incapable of capturing the long-term dependency, resulting in losing the important contextual information.
It has been shown that the RNN language models (RNNLMs) can be enhanced with some specific long-term contextual information, including document topics~\cite{mikolov2012context,ghosh2016contextual,dieng2017topicrnn}, bag-of-words contexts~\cite{wang2016larger-context}, a neural cache~\cite{grave2017imroving}, etc.
Several specific text structure was considered in the RNNLMs, such as the hierarchical sentence sequences~\cite{lin2015hierarchical}, tree-structured texts~\cite{tran2016inter} and dialog contexts~\cite{liu2017dialog,mei2017coherent}.

\begin{figure}[t]\centering
    {
    \includegraphics[width=.7\columnwidth]{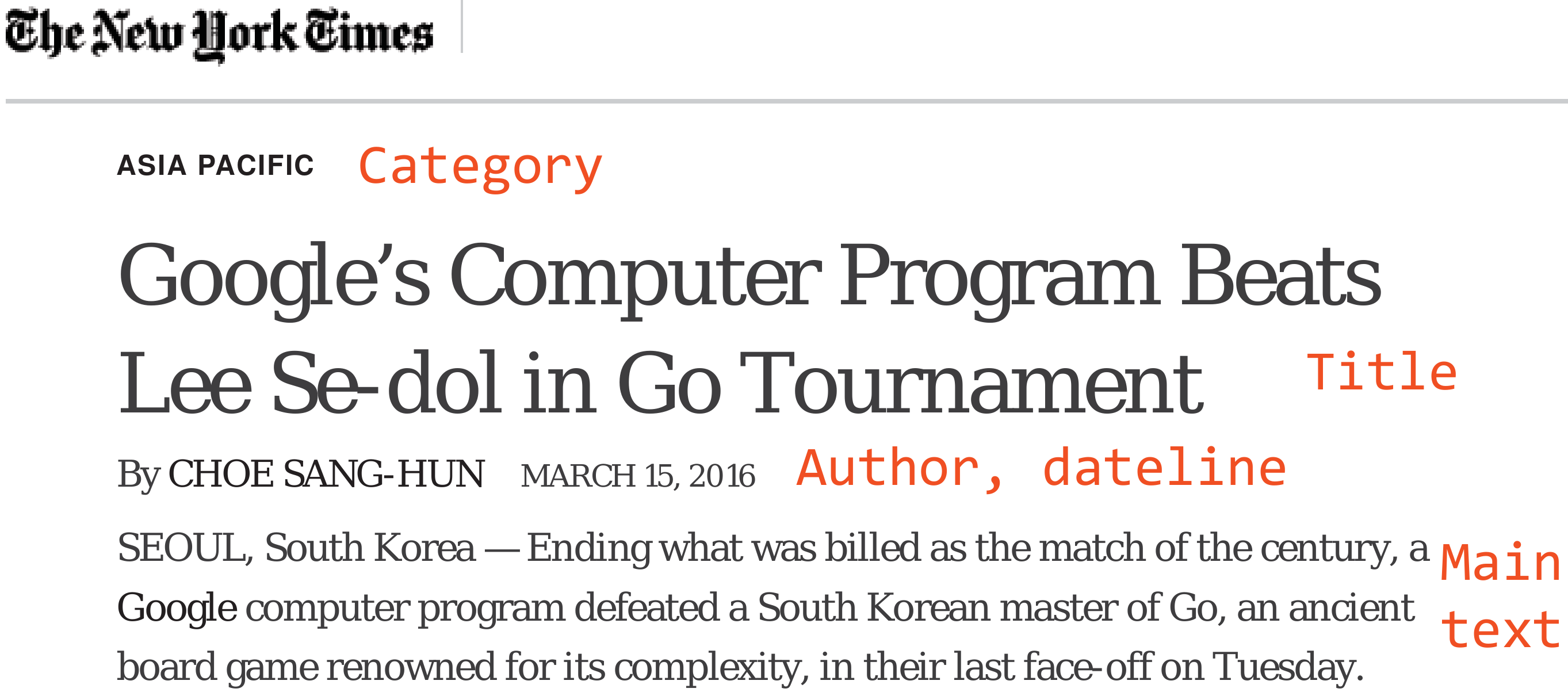}
    }
    \vspace{-.3cm}
    \caption{An AlphaGo News from the NY Times have several important semantic attributes, such as the title, the author, the category and the dateline.}
    \vspace{-.3cm}
    \label{fig:document modality}
\end{figure}

In the aforementioned models, only main text sequences were modeled but the vastly-accessible attributes of documents were ignored.
Interestingly, the document attributes implicitly convey global contextual information of the word distributions and are vastly-accessible before reading the main texts in daily reading or speaking. Document titles are compact abstracts carefully chosen by authors and keynote speakers. Labels and tags are specific categories assigned by experienced editors. Authorships reflect writing styles. With these vastly-accessible attributes, one can predict word distributions better (see a concrete example in Figure.~\ref{fig:document modality}).

Moreover, from the generation perspective, several previous works generate the designed outputs from scratch or from a single semantic attribute~\cite{lipton2015capturing,sennrich2016controlling,pavlick2016empirical,lebret2016neural,kiddon2016globally,radford2017learning,hu2017toward}. However, only a few semantic attributes were incorporated at the same time and were incapable to meet the huge complexity of the text generation task.
In this paper, we consider a diversity of semantic attributes and use the attention mechanism to conjoin the semantic attribute as a joint embedding.
Hence, the semantic attribute modulation brings a flexible way to generate texts because we can choose different combinations of these attributes. Due to the strong semantic information conveyed by the attributes, the text generations are \emph{interpretable} with regard to the different combinations of the input attributes.
With this flexibility, we can get a text style variation with replacements of semantic attributes. An interesting example is \emph{Please let Jason Mraz rewrite the lyric `Last Kiss'\footnote{A famous song by Taylor Swift.}}.

\subsection{Our Proposal}
In this paper, we present \method, the Semantic Attribute Modulation for language modeling and style variation.
We consider the vastly-accessible semantic language attributes and extract the attribute embedding.
Specifically, we adopt two types of semantic attributes: the title attribute and the category attribute. For the title attribute, we use an RNN encoder to get the title embedding. For the category embedding, our model learns a shared embedding from the documents in the specific category. Then, we generate the outputs with an attention mechanism over a diversity of attribute embeddings.


The semantic attribute modulated (\method) language model obtains better per-word prediction results than the vanilla RNNLM without \method. The improved word predictions are highly related to the semantic attributes and therefore \emph{interpretable} to humans. Moreover, we present the lyric generation task with lyric variation derived from semantic attributes. The text generation conditioned on the semantic attribute has a \emph{flexible} attribute selection.
With a learned attribute as a replaced input, we can get the output with the style variation. Interesting lyric style variations examples further demonstrate the \emph{flexibility} of \method.






In summary, our contributions are as follows:
\begin{itemize}
\item We present \method, a Semantic Attribute Modulation, which incorporates a diversity of semantic document attributes, as a \emph{flexible} language generation modulation input.
\item By incorporating the Semantic Attribute Modulation, our language model gets better word prediction results on several text datasets. The better word predictions are highly related to the semantic attribute and hence is \emph{interpretable} to humans.
\item Based on our model, we present the stylistic variations of the lyric generation with a fake author attribute, which further demonstrates the \emph{flexibility} of \method.
\end{itemize}

\section{Preliminaries}
In this section, we first give a concrete example of semantic attributes and then list the related language generation models.
\subsection{A concrete example of semantic attributes}\label{sec:semantic_attribute}
We take an AlphaGo news article from the New York Times
     as a concrete example~(Figure.~\ref{fig:document modality}).
Given the title `Google’s Computer Program Beats Lee Se-dol in Go Tournament', the main text words `Google', `program' and `Go' could be predicted more easily. Given the author attribute `CHOE SANG-HUN' who is a Pulitzer Prize-winning South Korean journalist, we can better predict the words `South-Korean' and `Go'. That is to say, the semantic attributes are indicative of different aspects of the text generation, which motivate us to modularize the semantic attributes in the text generation models.
\subsection{RNN-LM}\label{sec:rnnlm}
Given a sequence of words $x = (x_1, x_2, \cdots, x_n)$, language modeling aims at computing its probability $P(x)$ by
\begin{equation}
    P(\mathbf{x})=P(x_1,x_2,\cdots,x_n)=\prod_{i=1}^{n}P(x_i|x_ {<i}),
\end{equation}
where $x_{<i}$ are the words ahead of $x_{i}$.
We can use the recurrent neural network to build the word probabilities~\cite{mikolov2010recurrent}.
At each time step $i$, the transition gating function $\phi$ reads one word $x_i$ and updates the hidden state $\mathbf{h}_{i}$ as $
    \mathbf{h}_{i}=\phi(w_{i},\mathbf{h}_{i-1})$, where $w_{i}=E^{\top}x_{i}$ is the continuous vector representation of the one hot input vector $x_{i}$ and $E$ is the embedding matrix.
The probability of the next possible word $x^{*}$ in the vocabulary $V$ is computed by
\begin{equation}
    p(\hat{x}_{i+1}=x^{*})\propto\exp((\mathbf{W}_{h}\mathbf{h}_{i}+b)_{x^{*}}),
\end{equation}
where $\mathbf{W}_{h}\in\mathbb{R}^{|V|\times d}$, $b\in\mathbb{R}^{|V|}$ are the affine weights and biases respectively and $d$ is the dimension of the hidden state $\mathbf{h}_{i}$. Here the subscription $()_{x^{*}}$ specifies the specific column.

The RNN models were always criticized for their lacking capacity of the long-term sequential dependence, resulting in an unsatisfactory performance on modeling contextual information. Several previous works tried to capture the contextual information using the previous contexts.
Let $f(x_{<i})$ be the contextual representation extracted from the contexts and the generation process of the RNNLM with $f(x_{<i})$ is
\begin{equation}
P(\mathbf{x})=P(x_1,x_2,\cdots,x_n)=\prod_{i=1}^{n}P(x_i|x_ {<i},f(x_{<i})).
\end{equation}
The previous context representation $f({x_{<i}})$ can be extracted as the bag-of-words contexts~\cite{wang2016larger-context}, the latent topics~\cite{mikolov2012context} and the neural network embedding~\cite{ji2015document}.

\begin{figure*}[htbp]\centering
    {
    \includegraphics[width=.99\columnwidth]{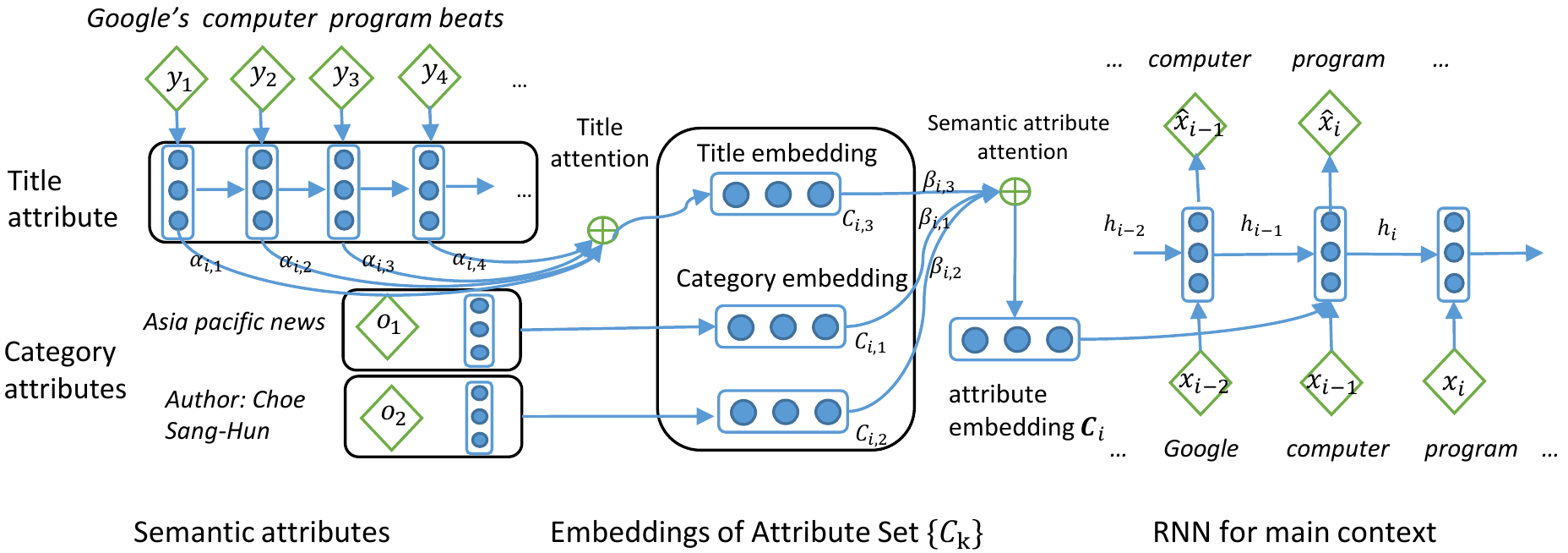}
    }
    \vspace{-0.1cm}
    \caption{The \method~architecture}
    \vspace{-0.3cm}
    \label{fig:model detail}
\end{figure*}
\section{Semantic Attribute Modulation}
Other than main texts, documents have semantic attributes, such as titles, authorships, tags, and sentiments, which convey important semantic information.
In this section, we present \method, the Semantic Attribute Modulation originated from an attention mechanism over a diversity of attributes. Then, we use \method~to do language modeling and style variation for language generation.
Given the semantic attribute modulated representation $\mathbf{C}$, the generative process of our model is $P(x_i|x_{0:i-1},\mathbf{C}), i=1,2,\cdots,n$, where $x_{i}$ are the words in the same document.

%
\subsection{Semantic Attributes}
Due to the discrepant forms among the semantic attributes, we use two methods to extract the representations from semantic attributes.

\subsubsection{Title Attributes}
The title is often carefully chosen by the author and is a compact abstract of a document.
Given an $m$-length title sequence $\mathbf{y}=(y_1,y_2,\cdots,y_{t},\cdots,y_m)$, we use a recurrent neural network to extract the hidden state $\vartheta_t$ of every title word $y_t$ as
\begin{equation}
\mathbf{\vartheta}_{t}=\phi(E^{\top}y_t,\mathbf{\vartheta}_{t-1}),
\end{equation}
where the dimension of the title word hidden state is $\tilde{d}$.
Since the title words do not have equal contribution to the whole context embedding, we use an attention mechanism for the title attribute, and obtain the different title representation $c_{i}$ for different main text words $x_{i}$ as a weighted sum:
\begin{equation}
 C_{i}=\sum_{t=1}^{m}\alpha_{t,i}\mathbf{\vartheta}_{t},
\end{equation}
where $\alpha_{t,i}$ is the attention value of the title word $y_{t}$ for the main text word $x_{i}$,
\begin{equation}
\alpha_{t,i}=\frac{\exp(a(\mathbf{\vartheta}_{t},\mathbf{h}_{i-1}))}{\sum_{t=1}^{m}\exp(a(\mathbf{\vartheta}_{t},\mathbf{h}_{i-1}))},
\end{equation}
$\mathbf{h}_{i-1}$ is the hidden state of the previous time step in the main text and $a$ is an attention function which scores how the title word $y_{t}$ affects the main text word $x_{i}$:
\begin{equation}
a(\mathbf{\vartheta}_{t},\mathbf{h}_{i-1})=\mathbf{\vartheta}_{t}M_{1}\mathbf{h}_{i-1}.
\end{equation}
With this title attention, we automatically learn different importance weights of the title words for each main text word.

\subsubsection{Category Attributes}
Category attributes are commonly used in daily writing and speaking. Useful category attributes include document categories, authorships, sentiments, etc. We formulate the category attribute as a one hot vector $o_k$ and the embedding of the category attribute is counted via an encoder of the one hot vector
\begin{equation}
C_k=o_k\cdot e_k,
\end{equation}
 where $e_k$ is a weight matrix which maps the one hot vector to a continuous category embedding.
We use the same embedding dimension for category attributes with the dimension of the title embedding as $\tilde{d}$.
\subsection{Language Generation and Style Variation with \method}
With the above semantic embedding extractions, we obtain a set of semantic attribute embeddings $\{C_{k}\}$.
To leverage the importance of each attribute for a main content word $x_{i}$, we adopt another semantic attribute attention mechanism to learn the semantic attribute embedding $\mathbf{C}_{i}$ for different main text words $x_{i}$ as
\begin{eqnarray}
\mathbf{C_{i}}&=&\sum_{k}\beta_{k,i}C_k,\\
\beta_{k,i}&=&\frac{\exp (b(C_{k},\mathbf{h}_{i-1}))}{\sum_{k}\exp (b(C_{k},\mathbf{h}_{i-1}))},
\end{eqnarray}
where $b(C_{k},\mathbf{h}_{i-1})=C_{k}M_{2}\mathbf{h}_{i-1}$ is an attention function which scores how the attribute $k$ affects the main text word $x_{i}$.

We incorporate the obtained semantic attributes into the RNN framework.
By using an attribute attention mechanism, the transition of RNN hidden state $\mathbf{h}_{i}$ reads not only the current word but also the semantic attribute embedding $\mathbf{C}_{i}$.
Specifically,we concatenate the semantic attribute embedding $\mathbf{C}_{i}$ and the input word embedding vector $E^{\top}x$. Thus, the hidden states update as:
\begin{equation}
\mathbf{h}_{i}=\phi(w_{i},\mathbf{h}_{i-1}), w_{i}=[E^{\top}x_i, \mathbf{C}_{i}].
\end{equation}

For the recurrent neural network function $\phi$, we use the gated recurrent unit~(GRU)~\cite{chung2014empirical}.
The GRU cell has two gates and a single memory cell. They are updated as:
\begin{eqnarray}\label{eqn:gru}
\nonumber\text{update gate: } z_{i}=\sigma(\mathbf{W}_{z}w_{i}+\mathbf{U}_{z}\mathbf{h}_{i-1})\\
\nonumber\text{reset gate: } r_{i}=\sigma(\mathbf{W}_{r}w_{i}+\mathbf{U}_{r}\mathbf{h}_{i-1}) \\
\nonumber\text{cell value: } \tilde{h}_{i}=\tanh(\mathbf{W}_{h}w_{i}+\mathbf{U}_{h}(\mathbf{h}_{i-1}\odot r_{i})) \\
\text{hidden value: }\mathbf{h}_{i}=(1-z_{i})\tilde{h}_{i}+z_{i}\mathbf{h}_{i-1}
\end{eqnarray}
where $\sigma$ is the sigmoid function and $\odot$ is the Hadamard product.
Our model is trained by maximizing the log-likelihood of the corpus, using the back-propagation through time~(BPTT) method~\cite{boden2002guide}.

As can be seen in Figure.~\ref{fig:model detail}, we build a Semantic attribute Modulated language generation model.
Semantic attributes can be considered as the inputs for the designed generation outputs.  
By comparing the semantic attributes, the corresponding outputs are \emph{interpretable} to users.
Moreover, considering that some attributes reflect the text styles, we realize the text style variation by replacing with some other related attributes. We will give some generated variations of the typical lyrics in the experiment part.

\begin{table*}[ht]
\caption{Statistics and Parameters of PTB, BBCNews, IMDB, TTNews, XLyrics}
\label{table:dataset-statistics}
\begin{center}
\setlength\tabcolsep{3pt}
\begin{tabular}{l|cccccc}
\hline\hline
& PTB  & BBCNews & IMDB & TTNews & XLyrics\\
 \hline
\#training docs&-&1,780&75k&70k&3.6k \\
\#training tokens& 923k&890k&21m&30m&118k\\
\#vocabulary&10k&10k &30k&40k&3k\\
attribute(s) &Category & Title+Category & Title& Author+title&Author+title \\
hidden size&200 & 1,000 & 1,000&1,000 & 1000 \\
\hline\hline
\end{tabular}
\end{center}
\end{table*}

\section{Discussions and Related Work}
\textbf{Neural Machine Translation}
Neural machine translation~(NMT) uses the encoder-decoder network to generate specific response~\cite{cho2014learning}. In NMT, the encoder network reads some source texts of one language and encodes them into continuous embeddings. Then the decoder network translates them into another language. NMT is also used to generate some poems after encoding some keywords~\cite{wang2016chinese}. This is similar to our work as generating some texts given some useful attributes. The difference from them is that our work uses a semantic attribute attention modulation to extract the semantic embedding instead of an encoder-decoder framework.

\textbf{Contextual RNN}    Our work is related to several contextual language modeling works.
In~\cite{hoang2016incorporating}, the titles and the keywords were represented as bag-of-words and used it to build a conditional RNNLM model. But this work only involved text attributes but could not model the discrete attributes. Discrete attributes, such as review rates and document categories, were also used to control the content generation~\cite{tang2016context}.
The variational auto-encoder based model with a generator-discriminator scheme was also used for generating controllable texts~\cite{hu2017toward} but the input attributes are limited to be only discrete categories.

There are several major advantages of our paper over the above methods. First, we adopt a more diverse attribute set, including the widely used category attributes. The semantic information brings the \emph{interpretability} of \method. Second, we use better attribute representation method, including a semantic attention mechanism and we can get \emph{flexibility} with the attention mechanism. Third, by replacing the semantic attributes, our model realize the style variation for the lyric generation.

\begin{table*}[ht]
\vspace{-0.2cm}
\caption{Word predictions that is improved, alike and worse after adding category attribute in Categories Politics, stocking and finance}
\label{table:word-probablity}
\begin{center}
\setlength\tabcolsep{3pt}
\begin{tabular}{l|c}
\hline\hline
&Words in the documents of the politics category  \\
\hline
Improved &  to, be, \textbf{ireland}, \textbf{bush}, one, \textbf{chairman}, \textbf{fiscal}, week, in, or, \textbf{plan}\\
Alike & both, general, many, both, but, is, N, in, been, the, said\\
Worse & of, gm, stock, orders, \textbf{law}, jerry \\
\hline
 \hline
&Words in the documents of the finance category  \\
\hline
Improved &  \textbf{exchange}, \textbf{share}, \textbf{group}, \textbf{third-quarter}, \textbf{soared}, from, is, \textbf{profit}\\
Alike &N,of, days, had, than, month, \textbf{share}, were, yield\\
Worse & reported, analysis, all, yield, vehicles, \textbf{economics}, gm, currently\\
 \hline
 \hline
\end{tabular}
\end{center}
\end{table*}
\vspace{-0.4cm}
\begin{figure*}[t]\centering
    {
    \includegraphics[width=.99\columnwidth]{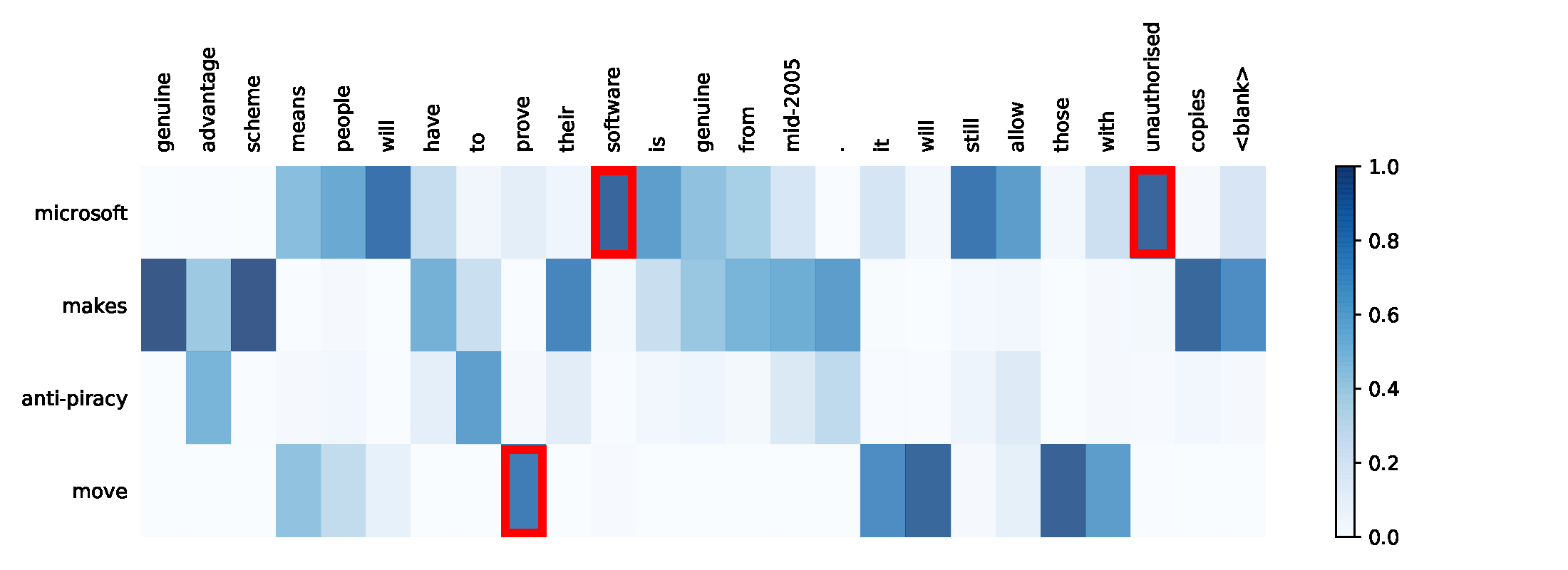}
    }
    \caption{An example of alignment matrix from \method-title (Best viewed in color)}
    \label{fig:attention}
\end{figure*}

\section{Experiments}
In this section, we first show that the Semantic Attribute Modulated language model gets better word predictions. The extensive qualitative analyses demonstrate the \emph{interpretability} of the word predictions with regard to the input attributes.  We then give several examples of the lyric style variation with \method, which shows the \emph{flexibility} of \method.
\subsection{Datasets}
We evaluate the proposed language model with semantic attribute attention on five different datasets with the different attribute combinations.
Among these datasets, TTNews, XLyrics and the titles of IMDB are collected by ourselves. We have the future plans to release the collected corpora after resolving the copyright issues. For detailed statistics, see Table.~\ref{table:dataset-statistics}.

\textbf{Penn TreeBank~(PTB)}~
Penn TreeBank~(PTB) is a commonly-used dataset for evaluating language models and its texts are derived from the Wall Street Journal~(WSJ). We use the preprocessed corpus by \citeauthor{mikolov2011extensions} and it has 929k training tokens with a vocabulary of size 10k~\footnote{\url{http://www.fit.vutbr.cz/~imikolov/rnnlm/simple-examples.tgz}}. We use the LDA topic model to analyze the PTB corpus with the topic number as $5$. We assign the label of one document as the topic assignment with the largest weight. The analysis of this category attribute and more discussions can be seen in Appendix A.

\textbf{BBCNews}
BBCNews is a formal English news dataset and contains 2,225 BBC news articles collected by \citeauthor{griffiths2004integrating}\footnote{\url{http://mlg.ucd.ie/datasets/bbc.html}}. The BBCNews documents have 5 class labels: business, entertainment, politics, sport and technology.

\textbf{IMDB Movie Reviews~(IMDB)}~
IMDB Movie Reviews~(IMDB) is a movie review corpus \cite{maas2011learning} and has 75k training reviews and 25k testing reviews\footnote{\url{http://ai.stanford.edu/˜amaas/data/
sentiment/}}. Note that \citeauthor{maas2011learning} did not provide review titles and we collected the titles according to the provided web links.

\textbf{TTNews}~
TTNews is a Chinese news dataset crawled from the several major Chinese media\footnote{\url{http://www.ywnews.cn/}, \url{http://www.toutiao.com}, \url{http://www.huanqiu.com/}, etc}. TTNews has 70,000 news articles with 30 million tokens and a vocabulary of size 40k. Each document contains the title and author annotations.

\textbf{XLyrics}~
XLyrics is a Chinese pop music lyric dataset crawled from the web. XLyrics has 4k lyrics, about 118k tokens and a vocabulary of size 3k\footnote{\url{http://www.xiami.com/song/1771862045?spm=a1z1s.6639577.471966477.105.3HI96A}}.

\subsection{Experimental Settings}
We consider several variants of the proposed methods with different combinations of semantic attributes. In detail, we consider the language modeling with a) a category attribute, b) a title attribute and c) a title attribute plus a category attribute. In order to realize the style variation of the generations, we consider generating lyrics with an original title attribute and a fake author attribute.

We train a recurrent language model without any side information as a baseline method. We also report the results of a count-based $n$-gram model with the Kneser-Ney smooth method~\cite{chen1996empirical,heafield2011kenlm}.


For training, we use the ADAM method with the initial learning rate of 0.001~\cite{kingma2014adam} to maximize the log-likelihood and use early-stop method based on the validation log-likelihood. The dimension of word embedding is set to be the same as the hidden size of RNN. The detailed parameter settings for each dataset are listed in Table.~\ref{table:dataset-statistics}.
\begin{table}[ht]
\caption{Corpus-level perplexity with Category-Attribute on (a) Penn Tree Bank and (b) BBCNews}
\label{table:language-modeling-category}
\begin{center}
\setlength\tabcolsep{3pt}
\begin{tabular}{l|cccc}
\hline\hline
& PTB & BBCNews\\
 \hline
5-Gram&141.2&131.1\\
RNN&117.1 &76.7\\
\method-Cat&\textbf{113.5}&\textbf{73.8}\\

 \hline\hline
\end{tabular}
\end{center}
\vspace{-0.2cm}
\end{table}
\begin{table*}[ht]
\vspace{-.2cm}
\caption{Corpus-level perplexity with Title-Attribute on (a) BBCNews and (b) IMDB}
\vspace{-.1cm}
\label{table:language-modeling-title}
\begin{center}
\setlength\tabcolsep{3pt}
\begin{tabular}{l|l|cccc}
\hline\hline
Attributes Source&Method& BBCNews & IMDB & TTNews& XLyrics\\
 \hline
\multirow{2}{1in}{Main Texts Only}&5-Gram&131.1&124.6 &136.7 &8.13\\
&RNN&76.7 &62.6&120.1 &7.56\\
\hline
\multirow{4}{1in}{+Titles}&RNN-State &72.2&61.0 & 118.2&8.20\\
&RNN-BOW & 72.2& 61.8& 118.4&8.18\\
&\method-Title-Att &\textbf{71.3} &61.3 &118.3 &7.56\\
&\method-Title-Att-State &72.5&\textbf{60.9}&\textbf{118.1} &\textbf{7.23}\\
\hline
\multirow{2}{1in}{+Titles+Authors}&\method-Title-Au-Att&-&-&114.1&7.08\\
&\method-Title-State-Au-Att&-&-&\textbf{113.4}&\textbf{6.84}\\
 \hline\hline
\end{tabular}
\end{center}
\vspace{-.2cm}
\end{table*}

\subsection{Language Modeling Word predictions}
We first show that the Semantic Attribute Modulated language model gets better word predictions. Then we give some qualitative analysis to show the interpretability of \method.
\subsubsection{Language Modeling with Category-Attribute}
Document categories are indicative of the discoursed topics and therefore of the distribution over words.
We first consider applying language modeling with category attribute on two corpora, PTB and BBCNews. For the PTB dataset, we use the LDA topic model to analyze the semantic information and we set the category as the topic which has the largest weight in LDA for every document. The details of the PTB dataset pre-processing can be seen in Appendix A. For the BBCNews dataset, we use the news category labels provided as a discrete category attribute.


 In Table.~\ref{table:language-modeling-category}, 5-Gram represents the count-based 5-gram model~\cite{chen1996empirical}, RNN represents the conventional RNN model without any semantic attribute and \method-Cat is our \method~model with a category attribute. As can be seen in the results, by adding a semantic category attribute, SAM-Cat outperforms the baseline models by achieving lower perplexities.

\subsubsection{Language Modeling with Title-Attribute}
Document titles are carefully chosen by the authors to summarize the document contents and attract the attention of readers. In this part, we incorporate the title attribute to take advantage of the implicit word distribution represented by the title. We use four corpora for this task. BBCNews and TTNews are two formal published corpora, IMDB is a movie review corpus and XLyrics is a lyric corpus.

 We implement the 5-gram model and the conventional RNN model on the corpus without titles. RNN-State is the conventional RNNLM model with the title's last hidden state as initialization. This means the title is considered as the first sentence but is not included in the prediction of per-word perplexities. RNN-BOW is the conventional RNNLM model incorporated with a bag-of-words representation of the title at each time step, which is a re-implementation of \cite{hoang2016incorporating}.  The \method-Title-Att method is the \method~model with the title attribute and the attention mechanism. By adding the title's last hidden state to \method-Title-Att as initialization, we get the \method-Title-Att-State method.


We show the word prediction perplexity results in Table.~\ref{table:language-modeling-title}.
The RNN-based models, with the title embedding, has better perplexity results. Moreover, \method-Title is better than RNN-state
because the added title information would disappear after several nonlinear gating functions.
The attention-based title attribute performs better than the one without attention. This is because the attention mechanism provides the different importance weights for the title words.

Generally, our \method~model with title attribute performs better on BBCNews, compared with IMDB. We believe the result is caused by the different genres of these datasets.
In order to make our title attribute useful, titles should be able to convey refined summaries of documents. BBCNews, as a formal news corpus written by professional journalists, usually has titles with higher quality than IMDB corpus. 

\subsubsection{Language Modeling with Title-Author-Attribute}

In this part, we incorporate two different attributes, title, and author. We will demonstrate that these two attributes are complementary.

We use the semantic attribute attention to conjoin the two attributes and the suffix `Au' means that this method incorporates the author categorical attribute and maintains the method notations used in the previous part.
We show the word prediction perplexity results of several attribute combinations in Table.~\ref{table:language-modeling-title}.
For the TTNews and XLyrics datasets, we can see that incorporating both title and author attributes are better than the single one.
\begin{figure*}[t]\centering
\begin{center}
    {
    \vspace{-0.4cm}
    \includegraphics[width=.99\columnwidth]{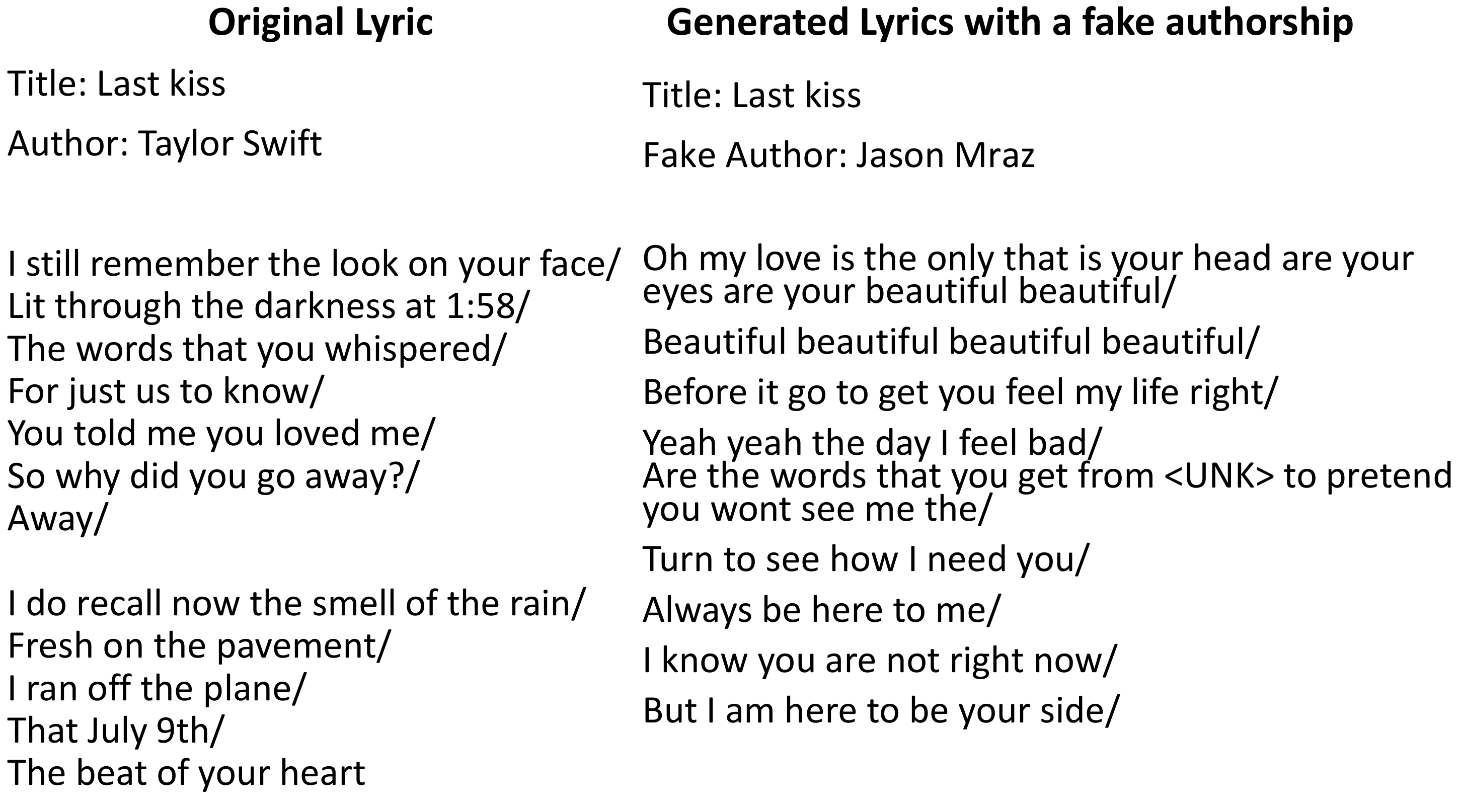}
    \vspace{-0.4cm}
    }
    \caption{Generated lyrics with the same title but a fake authorship. The original lyric is of the country style (left) and the generated lyric with a fake author is of the pop rock style (right).}
    \label{fig:lyric_en}
\end{center}
\end{figure*}
\begin{figure*}[t]\centering
\begin{center}
    {
    \vspace{-0.2cm}
    \includegraphics[width=.99\columnwidth]{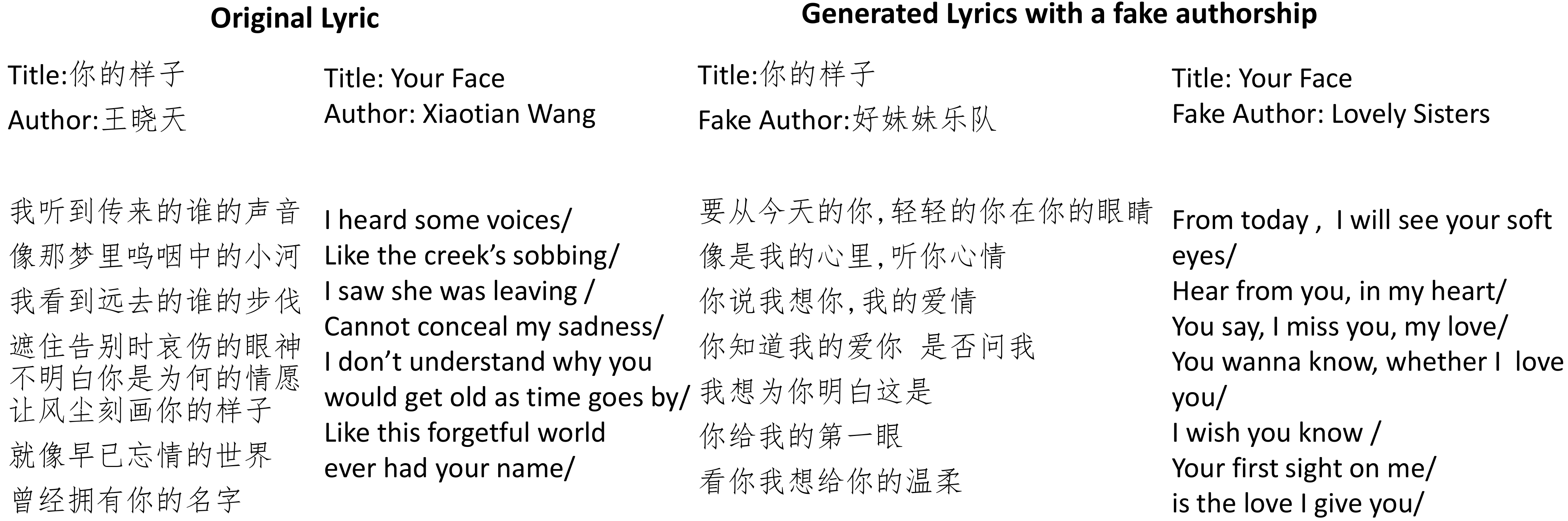}
    \vspace{-0.2cm}
    }
    \caption{Generated lyrics with the same title but a fake authorship. The original lyric is sentimental (left) and the generated lyric with a fake author is cheerful (right).}
    \vspace{-0.3cm}
    \label{fig:lyric_cn}
\end{center}
\vspace{-0.1cm}
\end{figure*}
\subsubsection{Qualitative Analysis on Interpretability of \method}
In order to discover why SAM-Cat outperforms traditional methods for the PTB dataset in Table.~\ref{table:language-modeling-category}, we demonstrate the words in the each category with the largest and the least perplexity changes in Table.~\ref{table:word-probablity}. We mark the words which have a strong semantic information of the each specific category in bold. For example for the politics category, after adding the category attribute, the words, which have the largest prediction improvement, are generally related to the politics, such as `Ireland', `bush' and `chairman'. The words, which have the largest prediction degeneration, generally have a semantic meaning but not related to the politics, such as `gm', `stock' and `orders'. The words, which have the least word prediction change, are generally function words, such as `both', `many' and `but'. The word prediction changes in other categories are similar with the politics category. We put the results of the finance category in Table.~\ref{table:word-probablity} and show the results of other categories in Appendix. B due to the space limit.


To further investigate how attention values control the importance weights of the attributes, we visualize some of the attention values in Figure.~\ref{fig:attention}. The color depth shows the attention weights. The red rectangles show the title word `Microsoft' has a large effect on the content words `software' and `unauthorized'. The title word `move' has a large effect on the content word `prove'. This example shows that the attention mechanism works as a \emph{flexible} selection of the attributes.
\subsection{Flexible Style Variation with \method}
Many downstream applications of the language modeling can be enhanced with the proposed semantic attributes. For machine translation, the semantic attributes could also be titles, authors, and categories. For the speech recognition task, the semantic attributes include the age and the dialect of the speaker. For language generation tasks, such as the question-answering and the poem/lyric generation, the possible attributes are titles, authors, and even styles.

We use the \method~model to perform lyric generation based and use both the title and author attributes.
Given an original lyric, we generate a new one with the same title but a fake author. We get several amazing generation results and the differences between two are highly related to the title attribute. Here we give two concrete examples (one in Chinese and the other in English) and left more examples in Appendix. C.

For the English example in Fig.~\ref{fig:lyric_en}: The original lyric \emph{last kiss} is a popular song by \emph{Taylor Swift} which is of the pop country style. After changing the authorship to \emph{Jason Mraz}, we generate a new love song which looks likes a rock lyric. The styles of the two lyrics tally the styles of the two singers.

For the Chinese example in Fig.~\ref{fig:lyric_cn}: The original lyric \emph{Your Face} is a sentimental love song written by \emph{Xiaotian Wang} which is recalling the past love. After changing the authorship to \emph{Lovely Sisters}, a trending Chinese band, we generate a joyful love song about the happiness of falling in love.


\section{Conclusion}
In this paper, we propose \method, the semantic attribute modulation for language modeling and style variation. The main idea is to take advantage of vastly-accessible and meaningful attributes to generate \emph{interpretable} texts. Our model adopts a diversity of semantic attributes including titles, authors, and categories. With the attention mechanism, our model automatically scores the attributes in a {flexible} way and embeds the attribute representations into the hidden feature spaces as the generation model inputs. 
The diversity of the input attributes make the model more powerful and interpretable and the semantic attribute attention mechanism brings flexibility for the whole model.
Extensive experimental demonstrates the effectiveness and the \emph{interpretability} of our flexible Semantic Attribute Modulated language generation model. 

In the future, we are interested in exploring more attributes which have semantic meaning for the language model task. In addition to the lyric generation task, other language generation tasks can also use our \method~model to utilize more semantic attributes. One possible example is to incorporate the geographic position attribute into the speech recognition task to model the dialects.
\clearpage
\bibliographystyle{alpha}
\bibliography{ref}
\clearpage
\begin{center}
\textbf{\huge\method: Semantic Attribute Modulation for \\Language Modeling and Style Variation} \\
\textbf{\huge (Appendix)}
\end{center}

\section*{Appendix A: Data Preparation of PTB}
PTB is a commonly-used corpus benchmark for the language modeling task. We use the LDA topic model to extract semantic category attributes.
Actually, adding a pseudo-category seems to be subtle for the language modeling task to see the words in advance and then predict them. We argue that the pseudo-category makes sense in the language modeling task evaluation for the following two reasons. First, We only add one discrete assignment for each document and there's no straightforward word distribution information propagated. Second, in fact, the category assignments have strong semantic information and we can find real category assignments for other datasets. The semantic analysis is as follows.

For the PTB dataset, we set the topic number as $5$ and set the largest topic weight assignment as each document's category assignment. As can be seen in Table.~\ref{table:topwords-ptb}, the topic $\#0$ focuses on the corporate finance, the topic $\#2$ focuses on the politics, the topic $\#2$ focuses on the managers, the topic $\#3$ focuses on the stocking market and the topic $\#4$ focuses on the daily news.

\begin{table*}[ht]
\caption{Top words of $5$ topics extracted from the PTB dataset}
\label{table:topwords-ptb}
\begin{center}
\setlength\tabcolsep{3pt}
\begin{tabular}{l|c}
\hline\hline
Topic& Top words\\
 \hline
0 & million billion share year company cents stock sales income revenue bonds profit corp.\\
1 & its mr. federal company u.s. new government state court plan officials bill house \\
2 & market stock trading prices stocks investors new price big index friday rates markets traders \\
3 & its company mr. inc. new co. corp. president chief executive says group chairman business vice \\
4 & mr. says when people years new time president work first few think good want city know back \\
 \hline\hline
\end{tabular}
\end{center}
\end{table*}

\section*{Appendix B: More word predictions of the SAM-Cat on PTB dataset}
In this part, we show some more word generations of our SAM-Cat model on the PTB dataset. We show that after adding the category attribute, we get more semantic word prediction improvements. In Appendix B, we show the results on the categories `stock' and `managements' in Table.~\ref{table:word-probablity-appendix}. We mark the words which have a strong semantic information of the each specific category in bold. After adding the category attribute, the words, which have the largest prediction improvement, are generally related to the category information. The words, which have the largest prediction degeneration, are generally have a semantic meaning but not related to the category information. The words, which have the least word prediction change, are generally function words.
\begin{table*}[ht]
\caption{Word predictions that is improved, alike and worse after adding category attribute in categories `corporate finance', `managers' and `stock market'}
\label{table:word-probablity-appendix}
\begin{center}
\setlength\tabcolsep{3pt}
\begin{tabular}{l|c}
\hline\hline
&Words in the documents of the stocking category\\
\hline
Improved &  \textbf{co}, \textbf{operating}, an, \textbf{markets}, considered, \textbf{commercial}, \textbf{stake}\\
Alike &N, usa, the is, these, discussion, at, the, chicken \\
Worse & offering, million, \textbf{money}, read, communications, lines, issues, city\\
 \hline
 \hline
&Words in the documents of the management category  \\
\hline
Improved &  \textbf{market},about,results,\textbf{orders}, \textbf{trading}, dow, \textbf{portfolio}, \textbf{price}, \textbf{market}\\
Alike &N, likely, of, prepared, southeast, futures, see, group, the\\
Worse & bear, totaled, optimistic, \textbf{executive}, chief, \textbf{manufacturers}, about\\
\hline
\hline
\end{tabular}
\end{center}
\end{table*}
\section*{Appendix C: More Lyric Generation Variations}
In this part, we give several lyric generation examples, two in English and one in Chinese. We observe that if the two authors have different style contents in the training data, the generation would very possibly be with different styles. In the following figures, we give the detailed generation and the corresponding analyses in the figure captions.

\begin{figure*}[t]\centering
\begin{center}
    {
    \vspace{-0.2cm}
    \includegraphics[width=.8\columnwidth]{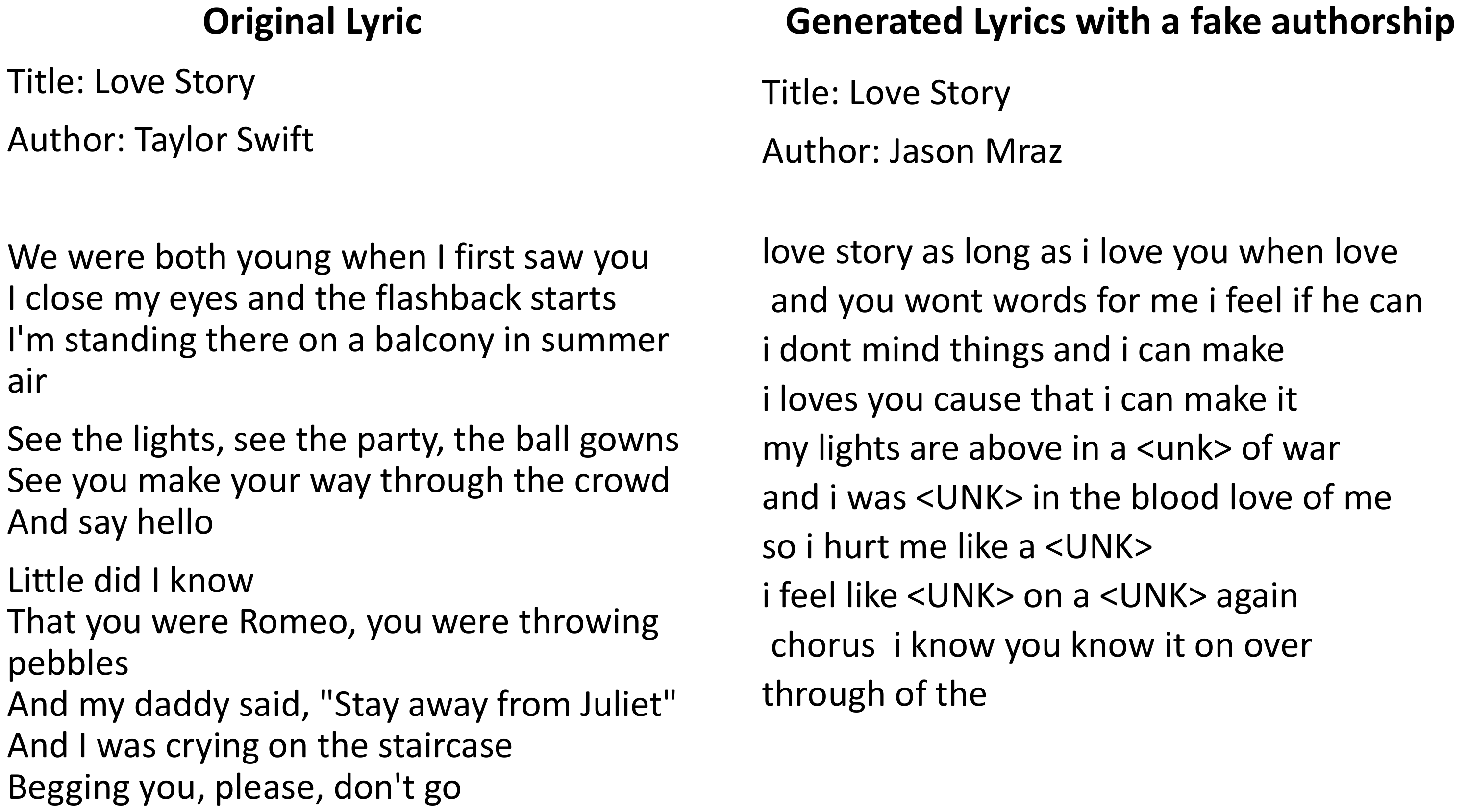}
    \vspace{-0.2cm}
    }
    \caption{Generated lyrics with the same title but a fake authorship. The original lyric is narrative (left) and the generated lyric with a fake author is whispering and piteous (right)}
    \vspace{-0.2cm}
\end{center}
\end{figure*}

\begin{figure*}[t]\centering
\begin{center}
    {
    \vspace{-0.2cm}
    \includegraphics[width=.8\columnwidth]{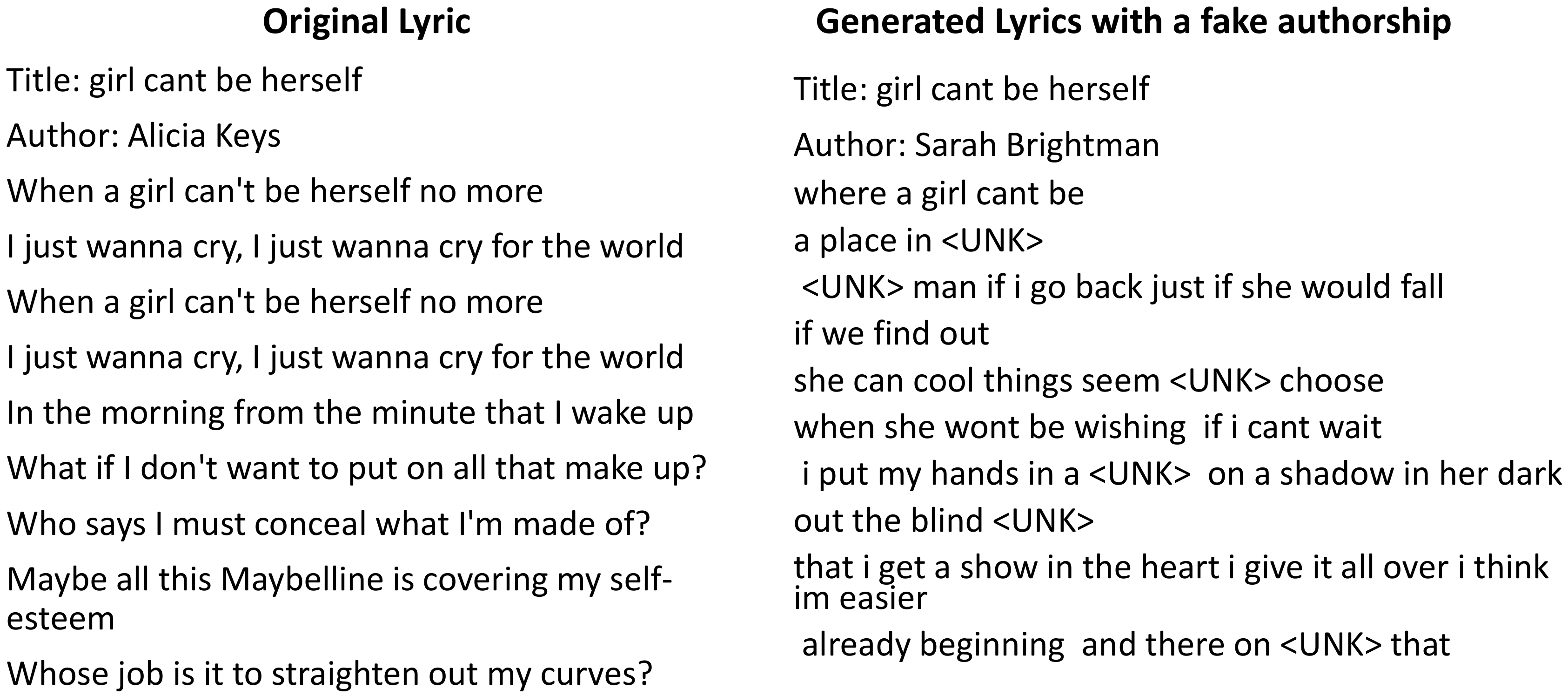}
    \vspace{-0.2cm}
    }
    \caption{Generated lyrics with the same title but a fake authorship. The original lyric is complaining (left) and the generated lyric with a fake author has a bystander view (right).}
    \vspace{-0.2cm}
\end{center}
\end{figure*}
\begin{figure*}[t]\centering
\begin{center}
    {
    \vspace{-0.2cm}
    \includegraphics[width=.8\columnwidth]{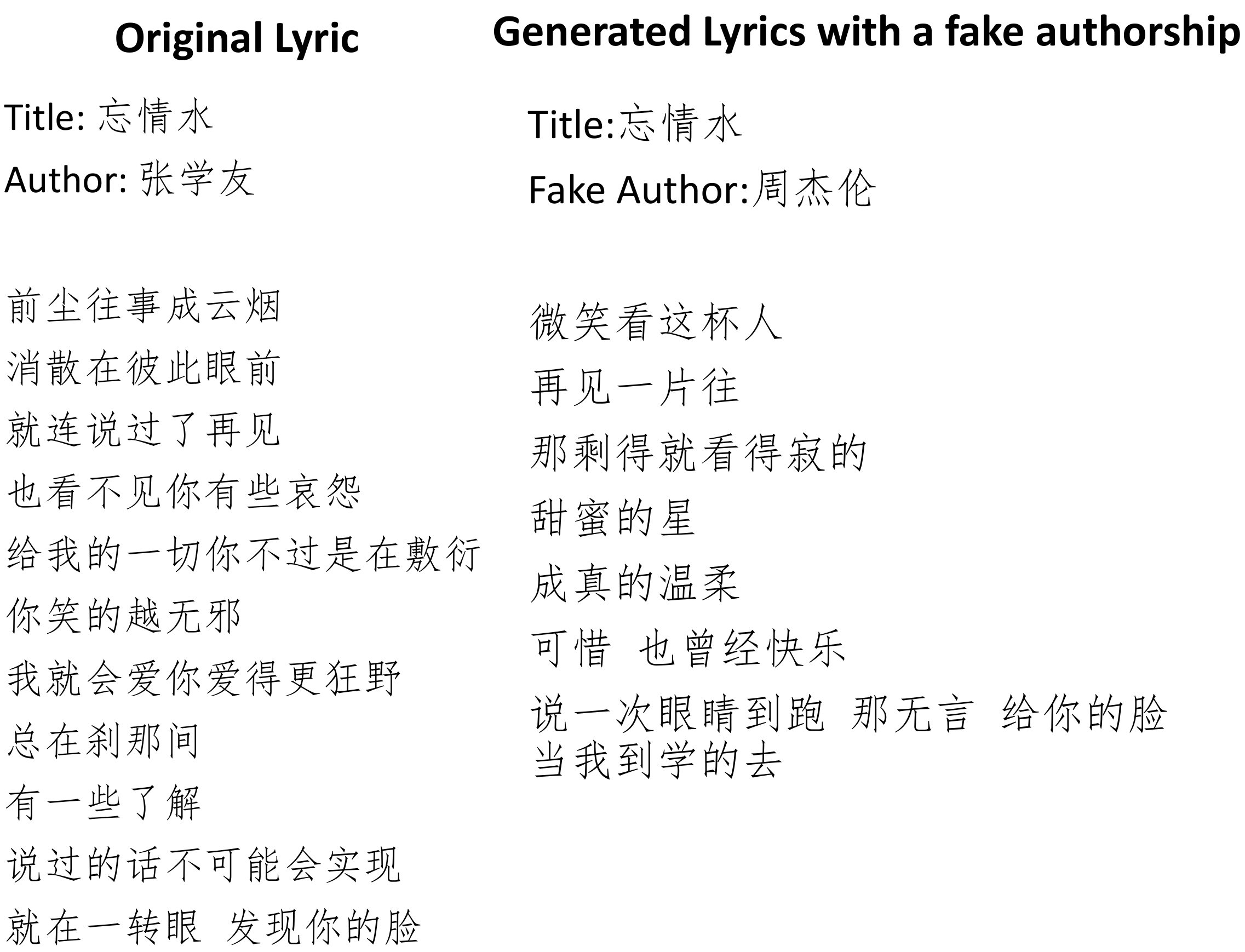}
    \vspace{-0.2cm}
    }
    \caption{Generated lyrics with the same title but a fake authorship. The original lyric is retro (left) and the generated lyric with a fake author is modern (right).}
    \vspace{-0.2cm}
\end{center}
\end{figure*}
\end{document}